\documentclass[letterpaper, 10 pt, conference]{ieeeconf}  

\IEEEoverridecommandlockouts                              

\usepackage{graphics} 
\usepackage{epsfig} 
\usepackage{amsmath} 
\usepackage{amssymb}  
\usepackage{hyperref}

\usepackage{booktabs}   

\usepackage[dvipsnames]{xcolor}

\title{\LARGE \bf
Hierarchical Reactive Grasping via Task-Space Velocity Fields and Joint-Space Quadratic Programming 
}

\author{
Anonymous Author(s)
}
\author{Yonghyeon Lee$^{1*}$, Tzu-Yuan Lin$^{1*}$, Alexander Alexiev$^{1}$, and Sangbae Kim$^{1}$
\thanks{*Equal contribution.}
\thanks{$^{1}$The authors are with the Department of Mechanical Engineering, Massachusetts Institute of Technology, Cambridge, MA 02139 USA (e-mail: yhl@mit.edu; tzuyuan@mit.edu; aalexiev@mit.edu; sangbae@mit.edu).}
}
 
\begin{document}

\maketitle
\thispagestyle{empty}
\pagestyle{empty}

\begin{abstract}
We present a fast and reactive grasping framework that combines task-space velocity fields with joint-space Quadratic Program (QP) in a hierarchical structure. Reactive, collision-free global motion planning is particularly challenging for high-DoF systems, as simultaneous increases in state dimensionality and planning horizon trigger a combinatorial explosion of the search space, making real-time planning intractable. To address this, we plan {\it globally} in a lower-dimensional task space -- such as fingertip positions -- and track {\it locally} in the full joint space while enforcing all constraints. 
This approach is realized by constructing velocity fields in multiple task-space coordinates (or, in some cases, a subset of joint coordinates) and solving a weighted joint-space QP to compute joint velocities that track these fields with appropriately assigned priorities.
Through simulation experiments and real-world tests using the recent pose-tracking algorithm FoundationPose~\cite{wen2024foundationpose}, we verify that our method enables high-DoF arm–hand systems to perform real-time, collision-free reaching motions while adapting to dynamic environments and external disturbances.
Project page: \href{https://reactivegrasp.github.io}{https://reactivegrasp.github.io}. 
\end{abstract}

\section{Introduction}

As robotic systems move toward unstructured and dynamic environments, the ability to adapt in real-time -- referred to as {\it reactive control} -- has become a key capability. Among core manipulation skills, grasping is one of the most fundamental and widely studied. Nonetheless, {\it reactive grasping} remains a challenging problem, particularly for high-dimensional systems, e.g., with multi-DoF grippers.

While recent work~\cite{liu2023target,fang2023anygrasp} has advanced real-time grasp-pose generation and tracking, the complementary challenge of collision-free reactive motion planning has received less attention. RL-based approaches~\cite{qin2023dexpoint,singh2024dextrah,huang2023earl} are often restricted to single-object, obstacle-free, top-down reaches, which limits their generalizability to out-of-distribution scenarios. In contrast, we propose a model-based real-time planning and control framework that generalizes naturally across clutter and diverse grasping directions.

Reactive grasping requires both real-time perception and motion planning, and we focus on the latter. 
We assume the availability of perception modules capable of recognizing object geometries and target fingertip positions, as well as dynamically tracking object poses -- assumptions that are becoming increasingly realistic thanks to recent advances in deep visual perception models~\cite{wen2024foundationpose,kim2022dsqnet,t2sqnet,wen2023bundlesdf,millane2024nvblox}.
In particular, we address the reaching phase -- the pre-grasp motion leading up to gripper closure -- formulated as a collision-free motion generation problem. Even with ideal perception modules, the high dimensionality of the arm–hand system, coupled with the complex and often highly concave geometries of everyday objects, makes real-time feedback motion generation a fundamentally challenging problem.

A substantial body of work addresses collision-free motion generation with closed-loop reactivity, relying on Artificial Potential Fields (APF)~\cite{khatib1986real, rimon1990exact, lacevic2013safety, paternain2017navigation} and, more recently, on Dynamical Systems (DS)~\cite{goncalves2010vector, khansari2012dynamical, huber2022avoiding, billard2022learning, huber2023avoidance}, where local modulation of a nominal motion plan enables efficient obstacle avoidance through analytical formulations.
Recent variants provide formal convergence guarantees for certain obstacle classes, including convex shapes, star-shaped objects, and tree-of-stars geometries. 

\begin{figure*}[!t]
    \centering
    \includegraphics[width=\linewidth]{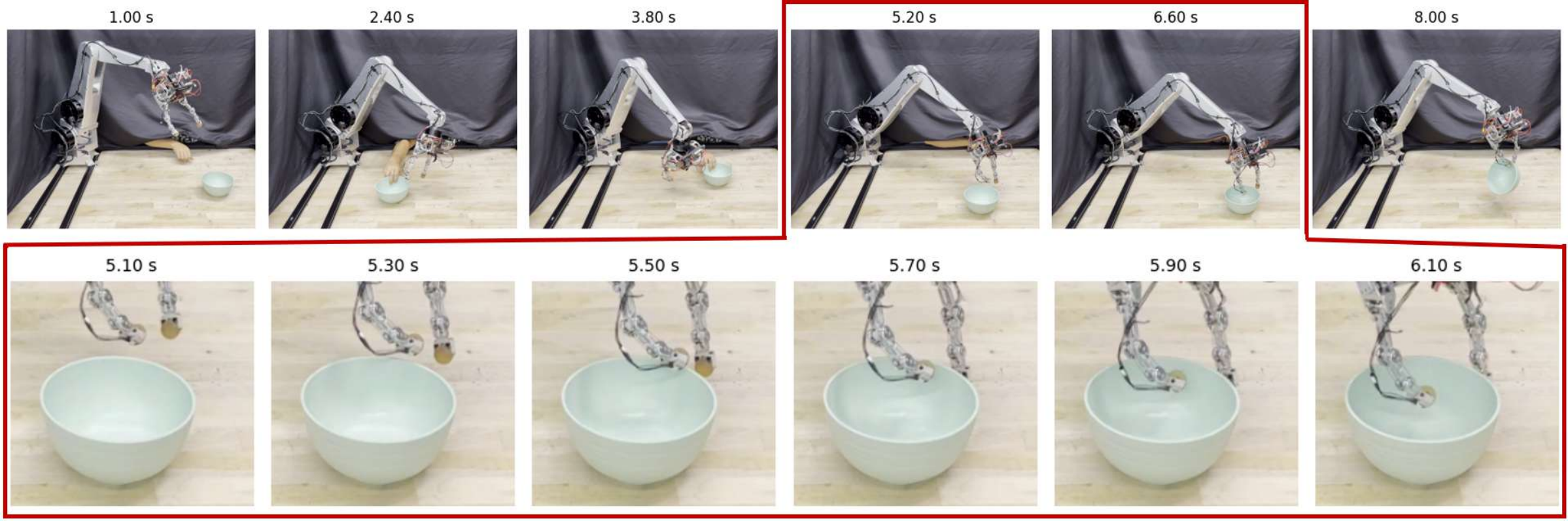}
    \caption{\emph{Reactive grasping} with a two-finger multi-DoF gripper (15-DoF arm–hand system: 7-DoF arm and 8-DoF hand; manipulation platform adapted from~\cite{saloutos2023towards}). {\it Upper}: Under dynamic disturbances to the target bowl position, the robot reactively approaches to grasp it. {\it Lower}: Near the bowl, the fingers curl inward to avoid collisions with the concave object and converge to a stable grasp pose.}
    \label{fig:examples}
    \vspace{-10pt}
\end{figure*}

For reactive grasping, these methods can be applied in either task space (e.g., at the level of individual links) or joint space, but both present challenges. 
In task space, multiple potential fields or dynamical systems are usually combined by weighting control forces projected into joint space via the Jacobian. 
This often results in (i) tedious weight tuning, (ii) poor representation of highly concave obstacles (e.g., wine glasses) when using structured shape abstractions such as tree-of-stars, and (iii) combined fields that lack convergence guarantees and are prone to local minima. 
In joint space, collision constraints are typically highly non-convex and unstructured, making them difficult to approximate with compact geometric abstractions and limiting the applicability of reactive modulation approaches.


Iterative motion optimization methods, such as Model Predictive Control (MPC), can generate joint-space, collision-free motions in a closed-loop manner and, with sufficiently long horizons, can escape local minima to achieve global planning~\cite{garcia1989model,williams2017information,williams2017model}. 
While traditionally too slow for global reactive planning with 7-DoF manipulators, recent advances -- such as GPU-accelerated solvers, geometric planners, and optimized multi-threaded CUDA implementations -- have significantly improved their practicality~\cite{bhardwaj2022storm,sundaralingam2023curobo}. Still, scaling to high-DoF arm–hand systems -- often more than twice the dimensionality of the arm alone -- remains difficult, as computational cost grows cubically with both state dimensionality and planning horizon. As a result, these methods are generally limited to quasi-reactive scenarios and require problem-specific sampling strategies or strong warm-start heuristics~\cite{domahidi2012efficient,richter2011computational,schulman2014motion,sacks2022learning,yoon2022sampling}.


In this paper, we present a highly reactive grasping framework for high-dimensional arm-hand systems that avoids the coupled growth of state and time-horizon dimensionality. The key idea is to plan {\it globally} in a lower-dimensional task space -- keeping the state dimension fixed and low -- with a subset of constraints, and to track {\it locally} in full joint space -- keeping the time horizon fixed and short -- while considering all constraints.
This is implemented as a two-layer hierarchical control framework.

In the higher layer, we construct time-invariant velocity fields -- first-order autonomous dynamical systems -- in multiple task-space coordinates, such as fingertip positions and orientations, as well as in subsets of joint coordinates, such as the gripper joints. These fields provide high-level velocity targets that guide the robot toward a desired grasp pose without accounting for complex joint-space constraints. 
In the lower layer, we solve a joint-space Quadratic Program (QP) that incorporates all relevant constraints -- including full-body collisions and joint velocity limits -- to compute feasible joint velocities that track these high-level targets with appropriate weighting.

At the heart of our framework is the principle that the high-level velocity targets are informed by the {\it global} paths, enabling them to guide the robot toward the target grasp pose. To this end, we construct linear fingertip velocities by solving lightweight global 3D path optimizations and extracting their initial velocity directions -- akin to the spirit of MPC, where only the first action is executed.
As a complementary design choice, we also incorporate auxiliary analytic velocity fields for fingertip orientations and gripper joints to further enhance grasp stability.

This hierarchical structure decouples global guidance from local feasibility: global planning operates in a fixed low-dimensional space with limited constraints, while local tracking is enforced through a one-step joint-space QP with full constraints, yielding a reactive strategy that is globally informed yet locally feasible.

We validate our framework through experiments in both simulation and the real world using a 15-DoF arm–hand system (7-DoF arm and 8-DoF hand; manipulation platform adapted from~\cite{saloutos2023towards}), demonstrating real-time collision-free reaching motion planning around concave target objects and multiple obstacles, as well as robust reactive grasping under disturbances to the robot and target objects; for example, Fig.~\ref{fig:examples} presents snapshots from real-world experiments.

\section{Methods}
We begin this section by introducing the assumptions and notations used throughout the paper. We consider a robotic manipulator equipped with a multi-DoF gripper comprising $m$ fingertips. The full arm-hand system has $n$ degrees of freedom and is fully actuated. The joint configuration is denoted by $q \in \mathbb{R}^{n}$, while the position and orientation of the $i$-th fingertip are denoted by $x_i \in \mathbb{R}^{3}$ and $R_i \in \mathbb{R}^{3 \times 3}$ for $i=1,\ldots, m$, respectively. 
We denote the concatenated vector of fingertip positions by $\mathbf{x} = (x_1, \ldots, x_m) \in \mathbb{R}^{3m}$.
The Jacobians associated with the $i$-th fingertip’s position and orientation are denoted by $J_i^x(q) \in \mathbb{R}^{3 \times n}$ and $J_i^R(q) \in \mathbb{R}^{3 \times n}$, respectively, such that
\begin{equation}
\dot{x}_i = J_i^x(q) \dot{q}, \quad \dot{R}_i R_i^\top = [J_i^R(q) \dot{q}],
\end{equation}
where $[{\omega}] \in \mathbb{R}^{3 \times 3}$ denotes the skew-symmetric matrix corresponding to the angular velocity vector $\omega \in \mathbb{R}^3$, such that $[\omega]v = \omega \times v$ for any $v \in \mathbb{R}^3$~\cite{lynch2017modern}.

We denote by $\Gamma(q) \in \mathbb{R}^{k}$ the vector of minimum Euclidean distances at configuration $q$ between each rigid link of the robot and the fixed environment (e.g., table, fixture, wall), as well as self-collision distances, where $k$ is the total number of considered collision pairs.
Ideally, $\Gamma(q)$ is differentiable everywhere; in practice, we assume that it is continuous and differentiable almost everywhere, and that the robot rarely passes through the non-differentiable points.

We assume that the geometry of the target object to grasp ${\cal O}$ is represented by a signed distance function $d(x, {\cal O})$, which outputs the signed Euclidean distance from a point $x$ to the surface of ${\cal O}$ -- negative inside the object, zero on the surface, and positive outside~\cite{lee2021imat}. The object is allowed to move dynamically, and a perception module is assumed to provide its time-varying signed distance function in real time, for example by tracking the object’s pose. 
With obstacles present, $d(x,\mathcal{O})$ can incorporate their signed distances, becoming vector-valued.

In addition, we assume that the perception module provides, in real time, a set of candidate fingertip target positions  
\[
\{ {\bf x}^*_j \in \mathbb{R}^{3m}\}_{j=1}^{N} \quad {\rm s.t.} \quad {\bf x}^{*}_j = (x^*_{j, 1}, \ldots, x^*_{j, m}),
\]
for the pre-grasp reaching phase (see Fig.~\ref{fig:dataset} for example). 
When these targets are predefined in the object frame, real-time generation reduces to tracking the object pose.

Given $d(x, \mathcal{O})$ and $\mathbf{x}_j^*$ in real time, we perform online reactive motion planning, executed at rates exceeding $50 \,\text{Hz}$ to ensure reactivity.
At each planning iteration, we select the target fingertip positions ${\bf x}^* \in \mathbb{R}^{3m}$ from the candidate set as those closest to the current positions ${\bf x}$, according to a suitable distance metric (this metric should be tailored to the specific gripper, and we provide an example in the experimental section). 
Then, our closed-loop planning algorithm for computing the desired joint velocities $\dot{q}_{\rm des} \in \mathbb{R}^{n}$ consists of the following three steps.

First, we solve a 3D collision-free path optimization for each fingertip, using the current positions ${\bf x}$ and selected targets ${\bf x}^*$ as boundary conditions. 
The optimization considers a subset of constraints, including the signed distances $d(x, \mathcal{O})$ to the target and non-target objects, as well as collisions with the table.
These global paths define local velocity fields for the fingertip positions; details are provided in Section~\ref{sec:fvf_via_3dpo}. 
Next, we construct auxiliary velocity fields for fingertip orientations and gripper joints to enhance grasp stability; see Section~\ref{sec:avf_for_gs}. 
Finally, the combined velocity fields are locally tracked by solving a joint-space Quadratic Program (QP) that computes feasible joint velocities under full-body collision and joint limit constraints; see Section~\ref{sec:wvft_via_qp}.
Implementation details are provided in Section~\ref{sec:id}.

\subsection{Fingertip Velocity Fields via 3D Path Optimization}
\label{sec:fvf_via_3dpo}
A straightforward approach to constructing a velocity field that drives the fingertips toward the target positions ${\bf x}^*$ is to use a linear attractor of the form
\begin{equation}    
\label{eq:linear_ft_vf}
\dot{{\bf x}}_{\rm des} = K({\bf x}^* - {\bf x}),
\end{equation}
where $K \in \mathbb{R}^{3m \times 3m}$ is a positive definite gain matrix, and ${\bf x} = (x_1, \ldots, x_m)$, ${\bf x}^* = (x_1^*, \ldots, x_m^*)$ denote the current and target fingertip positions, respectively~\cite{amanhoud2019dynamical}. However, when linear paths are obstructed by the target object itself -- even in the case of simple shapes such as cubes -- tracking this field under collision constraints can cause the robot to become trapped in local minima.

\begin{figure}[!t]
    \centering
    \includegraphics[width=\linewidth]{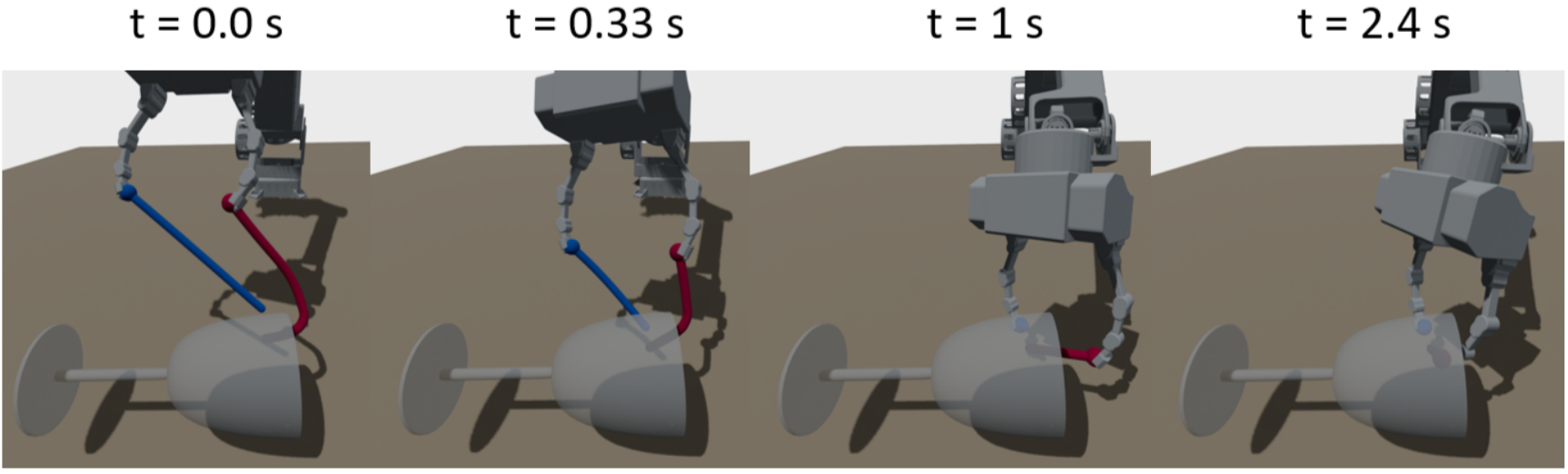}
    \caption{Real-time 3D path optimization for the fingertips.}
    \label{fig:3d_path_opt}
\end{figure}

To provide global guidance and mitigate local minima, we define a velocity field for each fingertip by solving a global path optimization problem in $\mathbb{R}^3$. Specifically, for the $i$-th fingertip position $x_i$, we solve
\begin{align}
\label{eq:3dpo}
    \min_{c_i(s)} \int_{0}^{1} \left\| \frac{d c_i(s)}{ds} \right\|^2 \, ds \quad \text{s.t.} &\quad d(c_i(s), \mathcal{O}) \geq \epsilon_{\mathcal{O}} \nonumber \\
    &\quad c_i(0) = x_i, \hspace{0.5em} c_i(1) = x^*_i
\end{align}
where $c_i(s)$ is a continuous path parameterized by $s \in [0,1]$, $\mathcal{O}$ is the target object, and $\epsilon_{\mathcal{O}} > 0$ is a safety margin to ensure clearance from the object surface.
When there are other non-target objects or obstacles in the scene (e.g., tissue boxes in Fig.~\ref{fig:bowl_two_obs_paths}), they can be incorporated as additional constraints using signed distance functions, with their geometry approximated by simple convex shapes that encapsulate them, such as superellipsoids~\cite{kim2022dsqnet}, since precise geometry is unnecessary. A relatively larger safety margin (e.g., 5 cm) can be used for non-target objects, whereas a tighter margin (e.g., 1 cm) is required for the target object. 
This 3D path optimization can be solved in real time for moderately complex objects using modern solvers~\cite{osqp, Andersson2019}; Fig.~\ref{fig:3d_path_opt} shows snapshots of the real-time fingertip path optimization results.

\begin{figure}[!t]
    \centering
    \includegraphics[width=\linewidth]{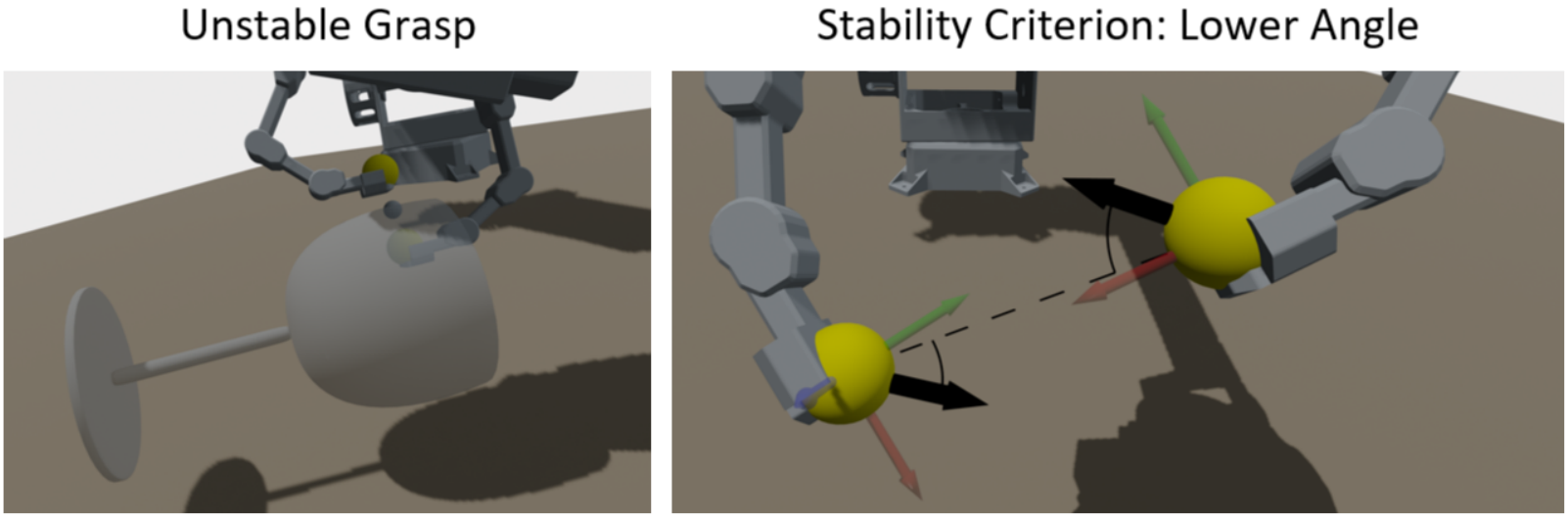}
    \caption{{\it Left}: An example of an unstable gripper configuration. {\it Right}: Stability score given by the angle between dotted line and black arrow.}
    \label{fig:two_fingers_stability}
    \vspace{-10pt}
\end{figure}

Let $x_i(s)$ be the solution of (\ref{eq:3dpo}), we then extract the initial velocity vector $\dot{c}_0 = \frac{d x_i}{ds}(0)$ from the optimized path and define the desired fingertip velocity as
\begin{equation}
\label{eq:ft_x_dot}
    \dot{x}_{i,\mathrm{des}} := v(x_i, x_i^*) \, \frac{\dot{c}_0}{\|\dot{c}_0\|},
\end{equation}
where $v(x_i, x_i^*) \in \mathbb{R}$ is a scalar speed function that depends on the current and target fingertip positions. To ensure a constant speed when the fingertip is far from the target and a smooth, quadratic decay to zero as it approaches the target, we define $v(x_i, x_i^*)$ as follows. Let $e_i := \|x_i - x_i^*\|$ denote the distance to the target. Then
\begin{equation}
\label{eq:csf}
    v(x_i, x_i^*) := 
    \begin{cases}
        v_{\mathrm{const}}, & \text{if } e_i > \delta_{x}, \\
        v_{\mathrm{const}} \left(1 - \left(1 - \frac{e_i}{\delta_{x}} \right)^2 \right), & \text{otherwise},
    \end{cases}
\end{equation}
where $v_{\mathrm{const}}$ is a constant maximum speed and $\delta_{x}$ is a threshold distance for speed modulation.

\subsection{Auxiliary Velocity Fields for Grasp Stability}
\label{sec:avf_for_gs}
Tracking only the linear fingertip velocities can drive the fingertips to their target positions, but may result in orientations that are unsuitable for stable grasping. For instance, as illustrated in Fig.~\ref{fig:two_fingers_stability} (\emph{Left}), in a two-finger gripper where each fingertip has a designated contact surface (yellow surface), the unintended part of the fingertip -- rather than the intended soft contact region -- may come into contact with the object, resulting in unstable or undesired grasps.

We assume that the stable grasping criterion is given as a scalar function of the fingertip positions and orientations, 
\begin{equation}
\label{eq:grasp_stability_orientation}
g(x_1, R_1, \ldots, x_m, R_m) \in \mathbb{R},    
\end{equation}
where the grasp is considered stable if \( g \leq \eta_g \), with $\eta_g > 0$ being a predefined stability threshold, and unstable otherwise. For example, in the case of a two-finger gripper shown in Fig.~\ref{fig:two_fingers_stability} ({\it Right}), the angle between the fingertip difference vector $x_2 - x_1$ (or $x_1 - x_2$) and the central axis of the desired contact region (illustrated as black arrows) -- computed from the fingertip orientation matrix $R_1$ (or $R_2$) -- can be used as the stability score $g$.

To guide the fingertip orientations toward satisfying the stable grasping criterion, we define an auxiliary angular velocity field for each fingertip as
\begin{equation}
\label{eq:des_omega}
\omega_{i, \mathrm{des}} = -{\rm unskew}\left( \frac{\partial g}{\partial R_i} R_i^\top \right),
\end{equation}
where ${\rm unskew}(\cdot) \in \mathbb{R}^{3}$ maps a $3 \times 3$ skew-symmetric matrix to its corresponding vector in $\mathbb{R}^3$ (i.e., the inverse of the skew operator $[\cdot]$) and $\frac{\partial g}{\partial R_i} \in \mathbb{R}^{3 \times 3}$ is a $3 \times 3$ gradient matrix.

Tracking the desired angular velocities in~(\ref{eq:des_omega}) becomes less important when the stability criterion is already satisfied, i.e., $g \leq \eta_g$, and should therefore be given lower priority in the tracking stage. To account for this, we introduce a weight parameter that depends on $g$:
\begin{equation}
\label{eq:weight_alpha}
    \alpha(g) :=  \frac{1}{2} + \frac{1}{2} \tanh\left(\frac{g - \eta_g}{T_g}\right),
\end{equation}
where $T_g$ is a temperature parameter that controls the smoothness of the transition. This function smoothly increases from 0 to 1 as $g$ increases past $\eta_g$, emphasizing orientation correction only when necessary.

Lastly, we introduce an auxiliary velocity field for the gripper posture, which drives the gripper toward a nominal configuration -- loosely open and ready to grasp -- when the current fingertip positions are far from the target. This nominal posture can often be easily predefined by the user. Specifically, let $q_{\mathrm{gr}}$ denote the current gripper joint value and $q^*_{\mathrm{gr}}$ the nominal gripper joint configuration. We define the desired gripper joint velocity as
\begin{equation}
\label{eq:q_gripper_dot_des}
    \dot{q}_{\mathrm{gr, des}} := v(q_{\mathrm{gr}}, q^*_{\mathrm{gr}}) \, \frac{q^*_{\mathrm{gr}} - q_{\mathrm{gr}}}{\|q^*_{\mathrm{gr}} - q_{\mathrm{gr}}\|},
\end{equation}
where $v(q_{\mathrm{gr}}, q^*_{\mathrm{gr}})$ is a scalar speed function, defined analogously to~(\ref{eq:csf}) with a threshold parameter $\delta_{\mathrm{gr}}$.

As with the fingertip angular velocity fields, this term should be downweighted when the fingertips are already close to their target positions. Denoting the fingertip position error by $h({\bf x}, {\bf x}^*) = \|{\bf x} - {\bf x}^*\|$, we define a weight parameter:
\begin{equation}
\label{eq:weight_beta}
    \beta(h) := \frac{1}{2} + \frac{1}{2} \tanh\left( \frac{h - \eta_h}{T_h} \right),
\end{equation}
where $\eta_h$ is a threshold distance and $T_h$ controls the smoothness of the transition. 

\subsection{Weighted Velocity Field Tracking via QP}
\label{sec:wvft_via_qp}
In this section, we present a QP formulation for tracking the desired velocity fields -- including the fingertip linear velocities $\dot{x}_{i, \mathrm{des}}$~(\ref{eq:ft_x_dot}), the fingertip angular velocities $\omega_{i, \mathrm{des}}$~(\ref{eq:des_omega}) weighted by $\alpha(g)$~(\ref{eq:weight_alpha}), and the gripper joint velocity $\dot{q}_{\mathrm{gr, des}}$~(\ref{eq:q_gripper_dot_des}) weighted by $\beta(h)$~(\ref{eq:weight_beta}) -- subject to full collision constraints, including $\Gamma(q)$ for robot-environment and self-collisions, and $d(x, \mathcal{O})$ for target object and obstacles avoidance.

We consider a constant desired joint velocity \( \dot{q}_{\mathrm{des}} \) over a planning horizon \( H > 0 \) -- which is a real scalar value --, and approximate the future joint configuration as \( q + \dot{q}_{\mathrm{des}} \, H\). Drawing inspiration from~\cite{mirrazavi2018unified, koptev2021real}, where a quadratic program is employed to solve inverse kinematics in real time, we formulate a QP that tracks the desired velocities by linearizing all constraints around the current configuration \( q \). A distinctive feature of our approach is the tracking of weighted velocity fields using position-dependent weights \( \alpha \) and \( \beta \):
\begin{align}
\label{eq:qp_formulation}
    \dot{q}_{\mathrm{des}} = \arg \min_{\dot{q}} \ \ & 
    \sum_{i=1}^{m} \left\| \dot{x}_{i, \mathrm{des}} - J_i^x(q) \dot{q} \right\|^2 \nonumber \\
    & + \alpha(g) \left\| \omega_{i, \mathrm{des}} - J_i^R(q) \dot{q} \right\|^2 \nonumber \\
    & + \beta(h) \left\| \dot{q}_{\mathrm{gr, des}} - S \dot{q} \right\|^2 \nonumber \\
    \noalign{\vskip 1.0ex}
    \text{subject to} \quad & \Gamma(q) + \frac{d\Gamma(q)}{dq} \, \dot{q} \, H \geq \epsilon_{\Gamma}, \nonumber \\ 
    & d(X(q), \mathcal{O}) + \frac{\partial d(X(q), \mathcal{O})}{\partial q} \, \dot{q} \, H \geq \epsilon_{\mathcal{O}}, \nonumber \\
    & q_{\rm max} \geq q + \dot{q} \, H  \geq q_{\rm min}, \nonumber \\
    & \dot{q}_{\rm max} \geq \dot{q} \geq \dot{q}_{\rm min},
\end{align}
where \( S \in \mathbb{R}^{n_{\mathrm{gr}} \times n} \) is a selection matrix that extracts the gripper joint components from the full joint velocity vector \( \dot{q} \in \mathbb{R}^n \), and \( n_{\mathrm{gr}} \) is the number of gripper joints. The term \( \Gamma(q) \in \mathbb{R}^k \) encodes robot-environment and self-collision distances, with threshold \( \epsilon_\Gamma > 0 \). The function \( d(X, \mathcal{O}) \in \mathbb{R}^l \) represents the signed distances between a set of points \( X \subset \mathbb{R}^{3l} \), approximating the gripper geometry, and the target object and non-target obstacles \( \mathcal{O} \), with safety margin \( \epsilon_{\mathcal{O}} > 0 \). Lastly, $q_{\mathrm{max}}$ and $q_{\mathrm{min}}$ denote the joint position limits, and $\dot{q}_{\mathrm{max}}$ and $\dot{q}_{\mathrm{min}}$ denote the joint velocity limits.

The above QP can be solved efficiently using analytical expressions for the Jacobians and the gradients of \( \Gamma(q) \) and \( d(X, \mathcal{O}) \). 
The planning horizon \( H \) and the margin parameters \( \epsilon \) control the conservativeness of the tracking behavior. 
In narrow passages, it may be necessary to reduce the desired velocity magnitude and choose a smaller planning horizon \( H \), along with tighter margins \( \epsilon \). 

\subsection{Implementation Details}
\label{sec:id}
This section provides implementation details. The specifications described here can be adapted to suit the user's specific application or system setup.
We use \texttt{Pinocchio}~\cite{carpentier2019pinocchio} for computing forward kinematics and Jacobians, and \texttt{FCL}~\cite{pan2012fcl} for computing distances and nearest points between collision geometries. These tools are used to evaluate $\Gamma(q)$ and its gradient. For convex objects, we employ superquadric models to represent the signed distance function~\cite{kim2022dsqnet}. 
For concave objects with thin surfaces (e.g., bowls, dishes, or the rim of a wine glass), we use a surface point cloud and define the signed distance as the nearest-point distance to this cloud. 
Given the thinness of the surface, ignoring the sign inside the object does not significantly affect the result.
As this computation can be expensive, we accelerate it using GPU-based nearest neighbor queries.

For the 3D path optimization~(\ref{eq:3dpo}), we discretize \( c_i(s) \) using 100 uniformly spaced grid points and solve the problem using Sequential Quadratic Programming (SQP) with a small number of iterations (typically \( \approx 3 \)).
For initialization, we use a GPU-accelerated sampling-based search to identify a single via point, positioned at a fixed distance (e.g., 10 cm) from the target, defining a piecewise linear path of two segments with minimal constraint violation.
We use \texttt{OSQP}~\cite{osqp} as the underlying solver for both the SQP used in the 3D path optimization and the joint-space QP~(\ref{eq:qp_formulation}). Given $\dot{q}_{\rm des}$ from (\ref{eq:qp_formulation}), we integrate over one control interval to obtain the desired joint position and then apply PD control with gravity compensation.
\section{Experiments}
We conduct case studies of the proposed reactive grasping method using a 15-DoF arm-hand manipulation system~\cite{saloutos2023towards}, consisting of a 7-DoF arm and two 4-DoF fingers (as shown in previous figures).
We first describe several design choices in our framework that are specific to this platform.
For the distance metric used to select the target fingertip positions -- i.e., to compare ${\bf x}=(x_1, x_2)$ and ${\bf x}^* = (x_1^*, x_2^*)$ -- we use
\begin{equation}
\label{eq:near_metric}
\|x_1 - x_1^*\| + \|x_2 - x_2^*\| + w \cdot \angle(x_2 - x_1,\; x_2^* - x_1^*),
\end{equation}
with a positive weight \( w \). 
The function $g(x_1, x_2, R_1, R_2)$ for the stable grasping criterion in (\ref{eq:grasp_stability_orientation}) is defined as the mean of the angles  
\[
\angle(x_2 - x_1,\; R_1^x + R_1^y) \quad \text{and} \quad \angle(x_1 - x_2,\; R_2^x + R_2^y),
\]  
where $R^x$ and $R^y$ denote the $x$- and $y$-axes of the orientation matrix, respectively, as illustrated in Fig.~\ref{fig:two_fingers_stability} ({\it Right}).

\subsection{Simulation Studies}
\begin{figure}[!t]
    \centering
    \includegraphics[width=0.9\linewidth]{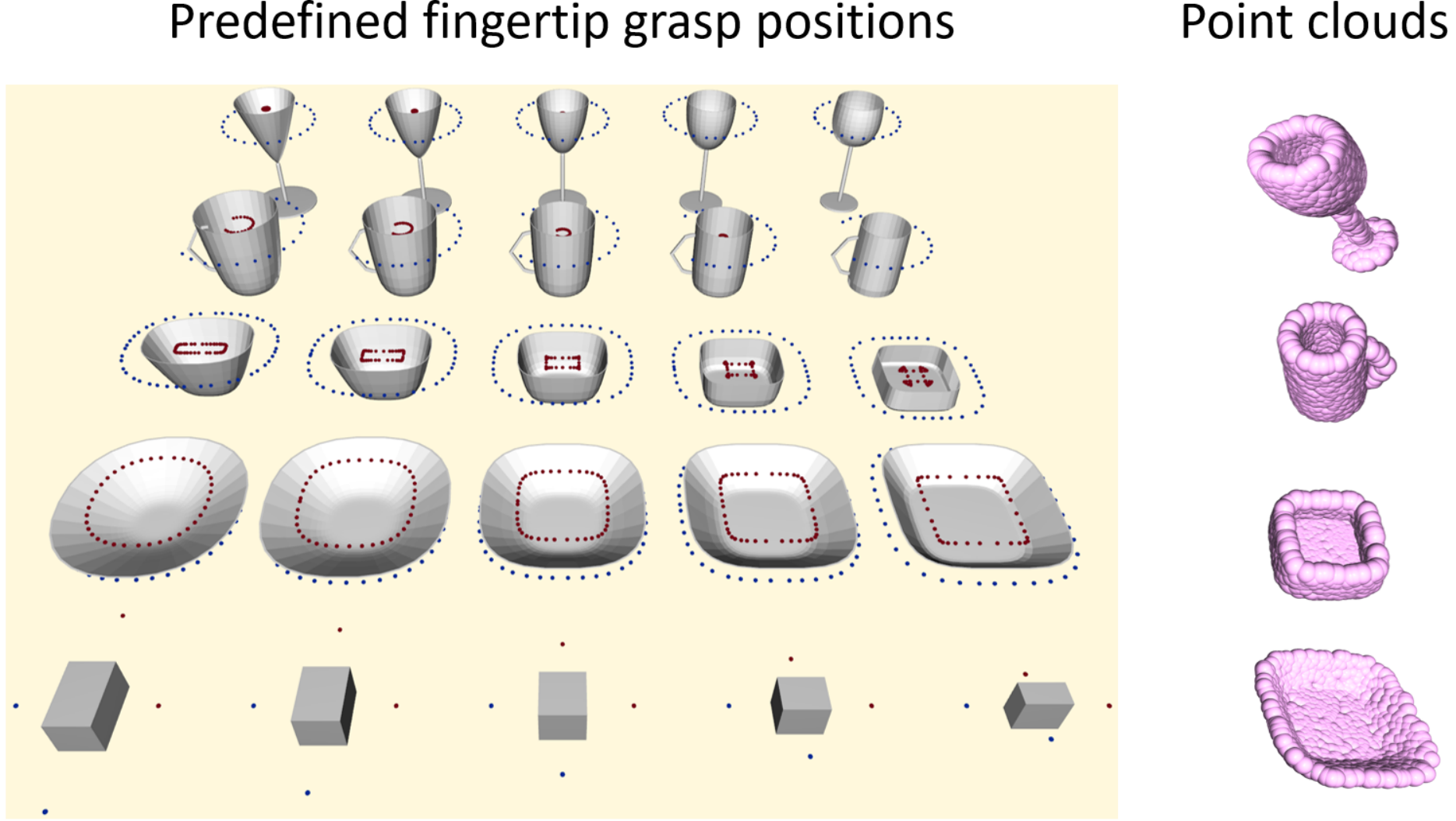}
    \caption{{\it Left}: A diverse set of objects, whose shapes and target fingertip positions are adopted from T$^2$SQNet~\cite{t2sqnet}. {\it Right}: Given point cloud representations of concave object surfaces, a distance threshold of $\epsilon_{\cal O} = 1 \ \mathrm{cm}$ yields collision geometries modeled as a set of spheres.}
    \label{fig:dataset}
\end{figure}

In this section, we present simulation studies in the Mujoco~\cite{todorov2012mujoco} physics engine using a diverse set of objects, as shown in Fig.~\ref{fig:dataset}. Each object is assumed to have predefined target fingertip positions, indicated by pairs of red and blue points; example collision geometries are shown in Fig.~\ref{fig:dataset} ({\it Right}).
The maximum planning frequency for $\dot{q}_{\rm des}$ is set to $100 \, {\rm Hz}$, running in a separate thread from the main control loop. Once the fingertips come within a specified threshold (1~cm) of the target positions, the gripper closes, the normal force is gradually increased until the object is firmly secured without slipping, and the object is lifted. 


\begin{figure}[!t]
  \centering
  \includegraphics[width=\linewidth]{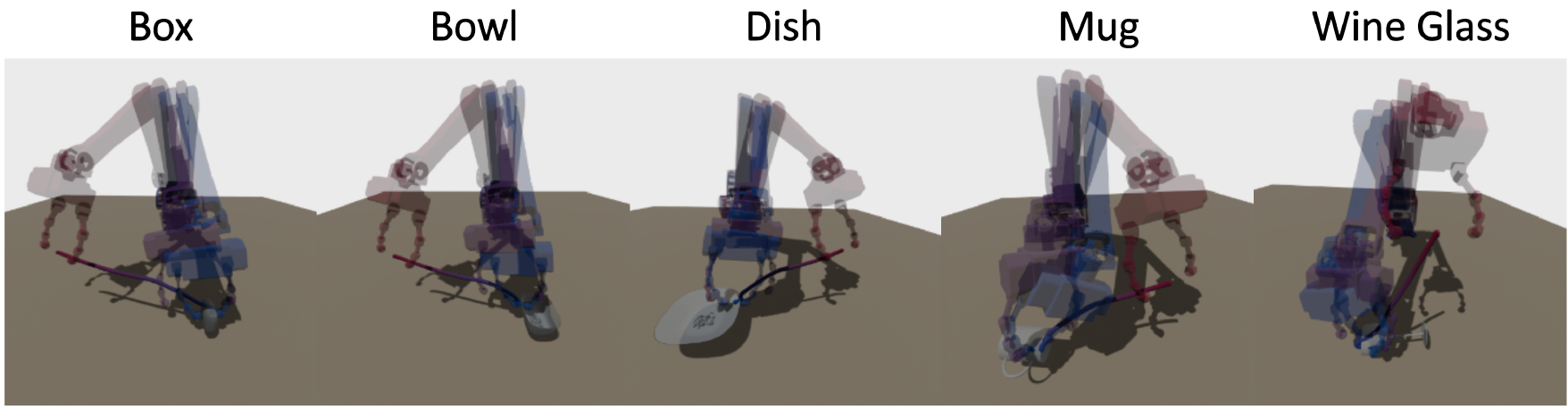}
  \caption{Example successful grasping trajectories using our method; robot motion progresses from red to blue.}
  \label{fig:results_ours_success}
  \vspace{-10pt}
\end{figure}

First, we demonstrate our algorithm’s ability to avoid local minima and successfully reach target positions for concave objects, in contrast to a simple linear fingertip velocity field in~(\ref{eq:linear_ft_vf}) adopted from~\cite{amanhoud2019dynamical}. 
Table~\ref{tab:grasp_success_rates} reports the grasp success rates.
The evaluation was conducted over three initial robot poses: gripper at left-bottom, center-up, and right-bottom.
For each pose, we tested five objects in each of five categories.
The Mug and Wine Glass were evaluated under five orientations (upright and rotated 90 degrees about each horizontal axis), while the other objects were tested only in the upright orientation.
An attempt is considered successful if the robot lifts the object within 20 seconds. The results clearly demonstrate the superiority of our method. Example successful grasping trajectories are visualized in Fig.~\ref{fig:results_ours_success}.

\begin{figure}[!t]
    \centering
    \includegraphics[width=\linewidth]{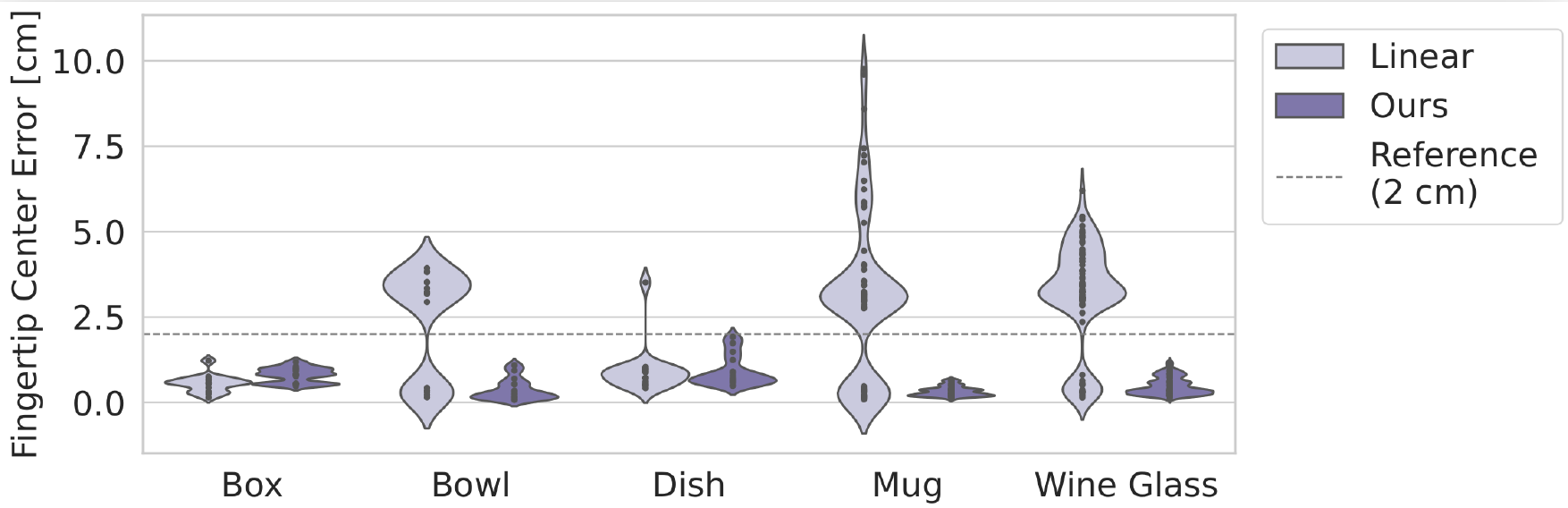}
    \caption{Distributions of fingertip center position errors at convergence, which must fall below approximately 2~cm to be considered successful.}
    \label{fig:ft_center_err}
\end{figure}

The causes of failure for the linear methods can be seen in Fig.~\ref{fig:ft_center_err}, which shows the fingertip position errors at convergence (i.e., at terminal time). As shown, the linear methods are frequently trapped in local minima, resulting in consistently large fingertip errors for all concave shapes. In particular, a closer look at the results for the Dish in Fig.~\ref{fig:ft_center_err} is interesting. For our method, the Dish exhibits slightly larger errors (up to $2 \, {\rm cm}$) compared to other objects (mostly below $1 \, {\rm cm}$). This is due to its thin shape, which brings a fingertip close to the table, where collision constraints slightly disturb the motion during gripper closing. For the linear methods, performance on the Dish is relatively better than on other concave objects, because its concavity is less pronounced. Although the linear paths result in collisions, iterative replanning helps avoid them.

\begin{table}[!t]
\centering
\small
\setlength{\tabcolsep}{3pt}
\caption{Grasp success rates across varying shapes, scales, and initial conditions.}
\label{tab:grasp_success_rates}
\begin{tabular}{lccccc}
\toprule
Method & Box & Bowl & Dish & Mug & Wine Glass \\
\midrule
Linear & 
100.0\% & 
33.3\% & 
93.3\% & 
30.7\% & 
32.0\% \\
Ours & 
{\bf 100.0\%} & 
{\bf 100.0\%} & 
{\bf 100.0\%} & 
{\bf 100.0\%} & 
{\bf 100.0\%} \\
\bottomrule
\end{tabular}
\end{table}

\begin{figure}[!t]
    \centering
    \includegraphics[width=1\linewidth]{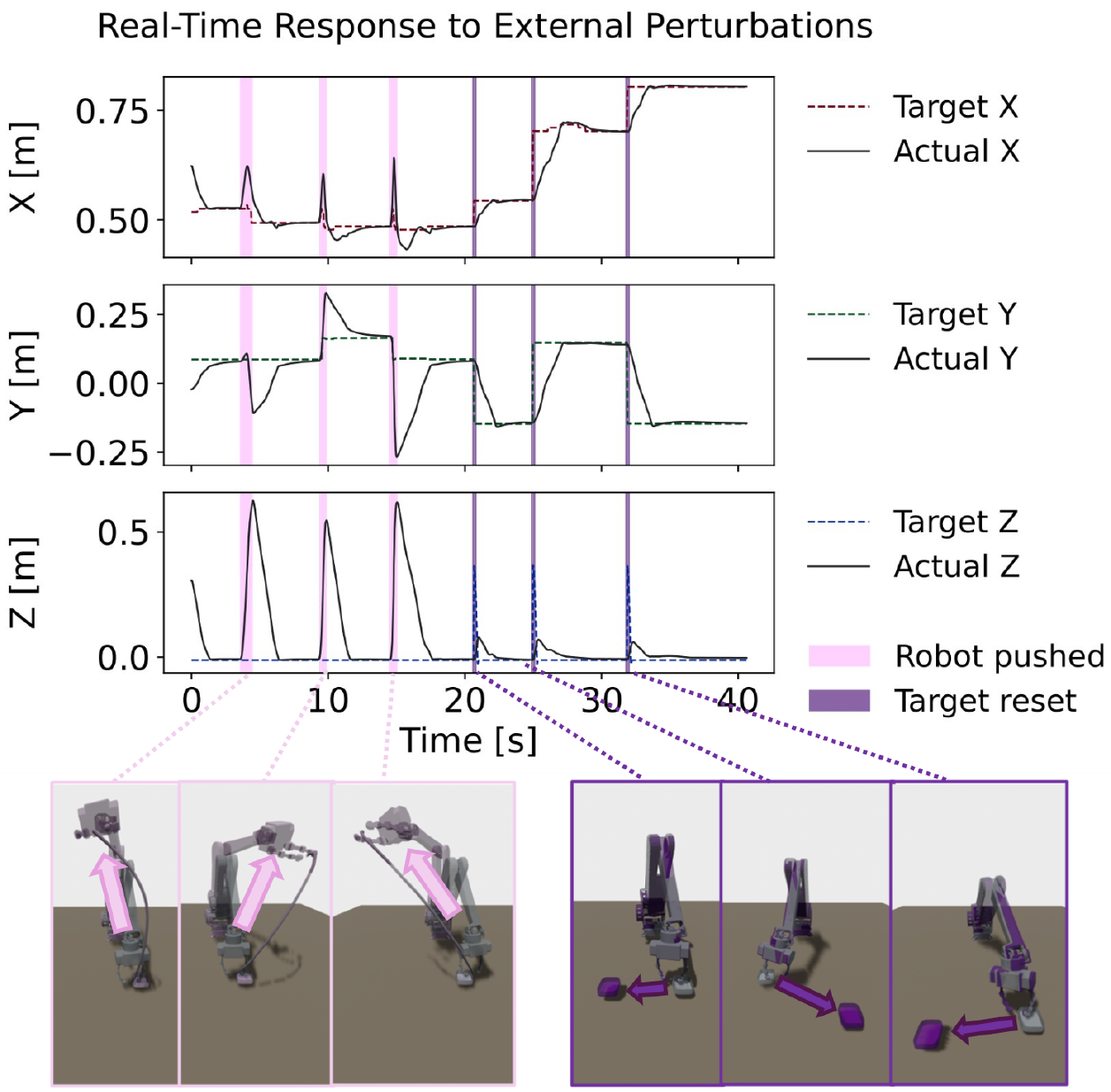}
    \caption{Real-time response of our method to external perturbations. Shaded pink regions indicate physical pushes to the robot, and shaded purple regions mark sudden changes in the target object's position. Dashed lines represent the target fingertip center coordinates $(x, y, z)$, while solid lines indicate the actual trajectories. }
    \label{fig:reactive_results}
\end{figure}

\begin{figure}[!t]
    \centering
    \includegraphics[width=\linewidth]{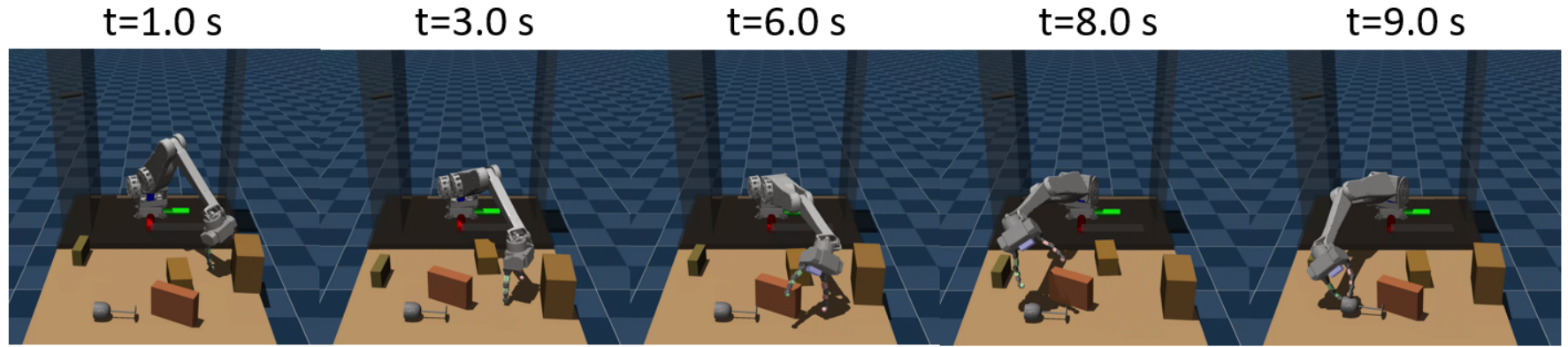}
    \caption{Reactive reaching toward the wine glass without colliding with the moving boxes.}
    \label{fig:multiple_obstacles}
    \vspace{-10pt}
\end{figure}


Next, we perform a perturbation analysis. Fig.~\ref{fig:reactive_results} illustrates the system’s real-time response to external disturbances, including physical pushes to the robot (marked in light pink) and sudden changes in the target object’s position (marked in dark purple), both triggered interactively in the physics engine. Dotted curves denote the target x-, y-, and z-coordinates of the fingertip center positions. While changes in the target position cause discontinuous jumps in these curves, slight changes also occur during robot pushes -- even when the object remains stationary -- due to updates in the nearest target point as the robot moves. Overall, the robot consistently recovers and re-converges to the target despite perturbations, demonstrating the real-time closed-loop reactivity of our grasping method. 

Lastly, Fig.~\ref{fig:multiple_obstacles} shows our method applied to a scene with multiple non-target moving objects that limit the collision-free area; note that we use superellipsoids for these objects, while the target is represented by a point cloud. Our reactive grasping successfully navigates around the four moving boxes and reaches the target wine glass without collision.

\subsection{Real Robot Experiments}
\begin{figure}[!t]
    \centering
    \includegraphics[width=1\linewidth]{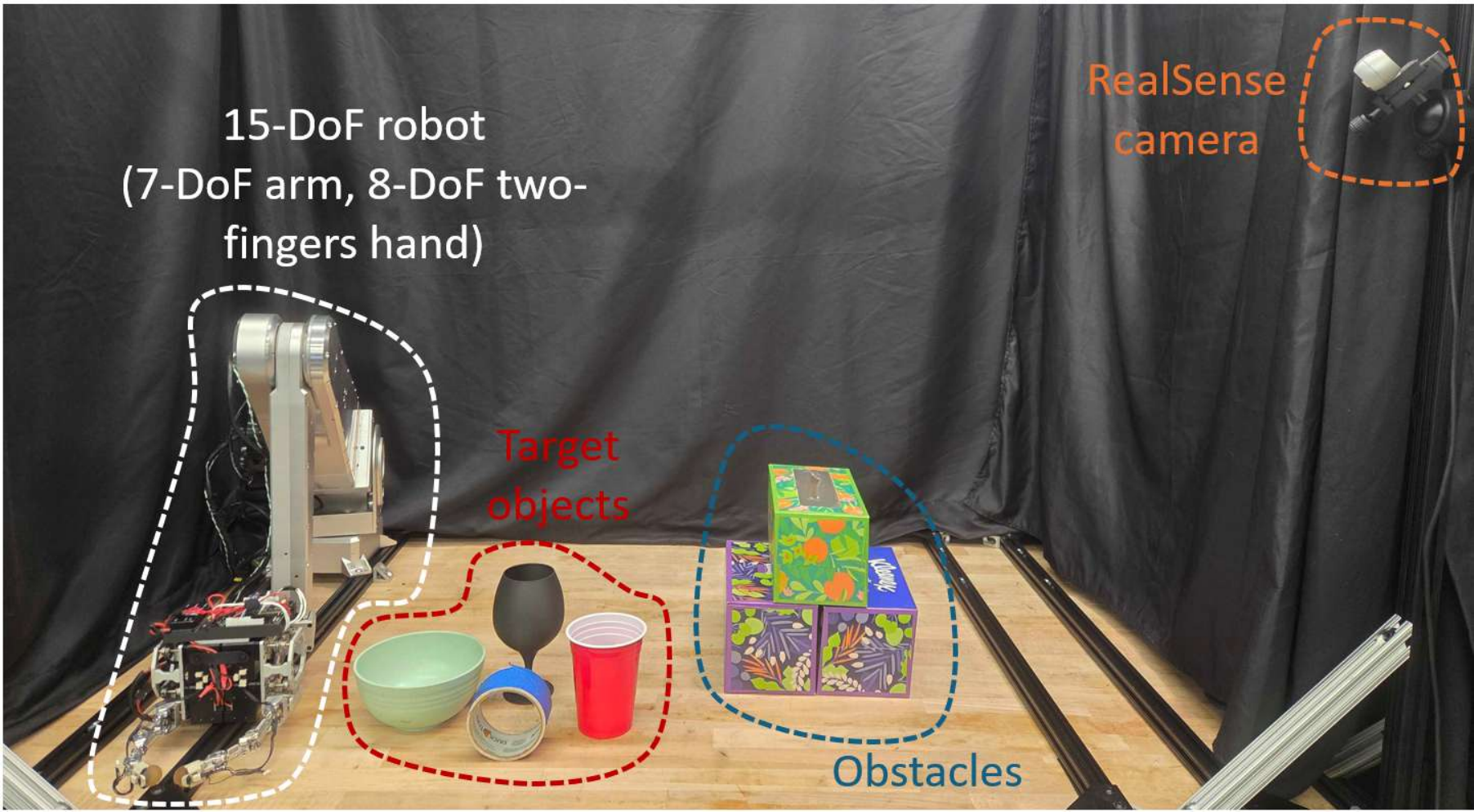}
    \caption{Experimental setup.}
    \vspace{-10pt}
    \label{fig:exp_setting}
\end{figure}

This section evaluates the proposed method in real-world settings, examining its ability to respond in real time to dynamic disturbances of the robot and target object, and to navigate through multiple obstacles. An overview of the experimental setup is shown in Fig.~\ref{fig:exp_setting}. We deploy the proposed algorithm on a 15-DoF robot manipulation system from~\cite{saloutos2023towards}. In total, four concave objects and three obstacles of the same geometry are used in the experiments. For perceptual measurements, we use an externally mounted Intel RealSense D455 camera. In all experiments, we consider collisions with the walls, table, obstacles (modeled as superellipsoids~\cite{t2sqnet}), target objects (represented as point clouds), and the robot itself.

Throughout the experiments, we assume prior knowledge of the target object model and its predefined grasp points. Object meshes are reconstructed with BundleSDF~\cite{wen2023bundlesdf} and tracked with FoundationPose~\cite{wen2024foundationpose} for real-time pose estimation. Grasp fingertip positions are manually specified, as in Fig.~\ref{fig:dataset} ({\it Left}). The robot is controlled at $500 \, {\rm Hz}$ with a separate planning thread at $100 \, {\rm Hz}$ (the actual planning speed can vary with factors such as the number of obstacles, but it generally remains above $50 \, {\rm Hz}$). As in the simulations, once the fingertips are within $1.5 \, {\rm cm}$ of their targets, the gripper closes and lifts the object.

An episode is deemed successful if the robot lifts the object. Collisions with the target object are not considered failures, while collisions with other obstacles are (except incidental contacts with unmodeled elements such as cables). Trials were excluded if failures were clearly due to hardware issues -- e.g., motor failures -- or if FoundationPose~\cite{wen2024foundationpose} completely lost the object pose, primarily from occlusions, to isolate algorithmic performance from mechanical limitations and perception failures.

\begin{figure}[!t]
    \centering
    \includegraphics[width=1\linewidth]{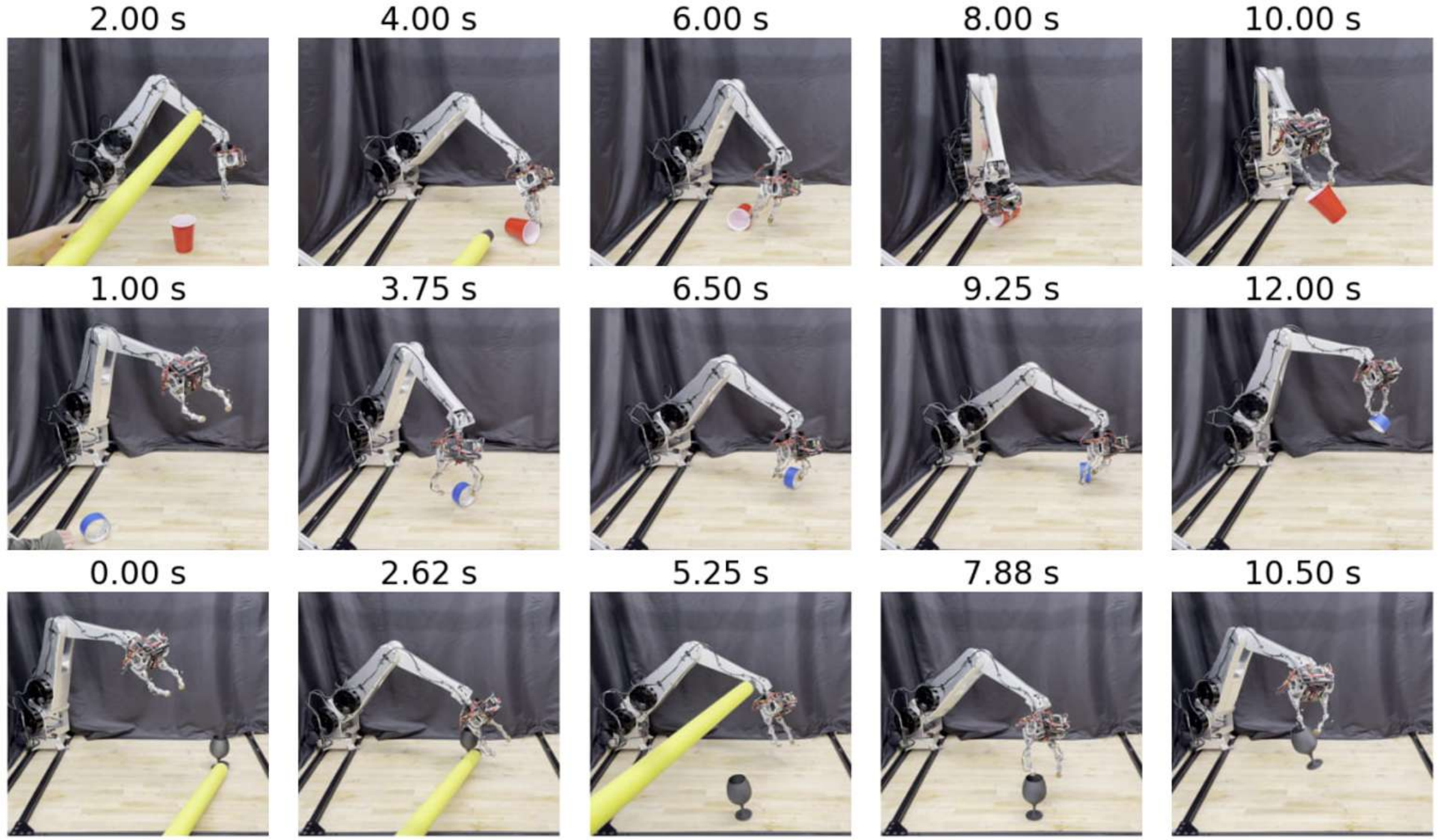}
    \caption{Examples of strong disturbances (large deviations in object position and strong pushes to the robot).}
    \label{fig:example_disturbances}
\end{figure}

First, we evaluate single-object grasping without additional obstacles, under strong disturbances applied either to the robot or to the object (see Fig.~\ref{fig:example_disturbances}). These disturbances are introduced after the robot has begun moving and occur multiple times per trial. Specifically, while conducting strictly controlled experiments is challenging, we constrain the disturbances to remain within reasonable bounds: (i) the objects must stay within the robot’s workspace, (ii) they must not be moved into ungraspable configurations (e.g., an upside-down bowl), and (iii) their motion must remain within the speed range that FoundationPose can reliably track (see supplementary videos for examples).
We performed five trials per object. The bowl, cup, and wine glass tasks were all successful, whereas the tape task resulted in one failure. Although the robot fingers occasionally collided with the target object due to tracking errors, the system was generally able to recover and complete the grasp. The failure with the tape object -- where the gripper closed prematurely while the tape was still rolling, resulting in a missed grasp -- was an expected outcome, as no prediction module for the object’s future position was used.

\begin{table}[!t]
\centering
\caption{Grasp success rates for the bowl with varying numbers of obstacles (successes out of 10 trials).}
\label{tab:success_rates_obstacles}
\begin{tabular}{c|ccc}
\toprule
$\#$ Obstacles & 1 & 2 & 3 \\
\midrule
 & 8/10 & 8/10 & 7/10 \\
\bottomrule
\end{tabular}
\vspace{-10pt}
\end{table}

\begin{figure}[!t]
    \centering
    \includegraphics[width=1\linewidth]{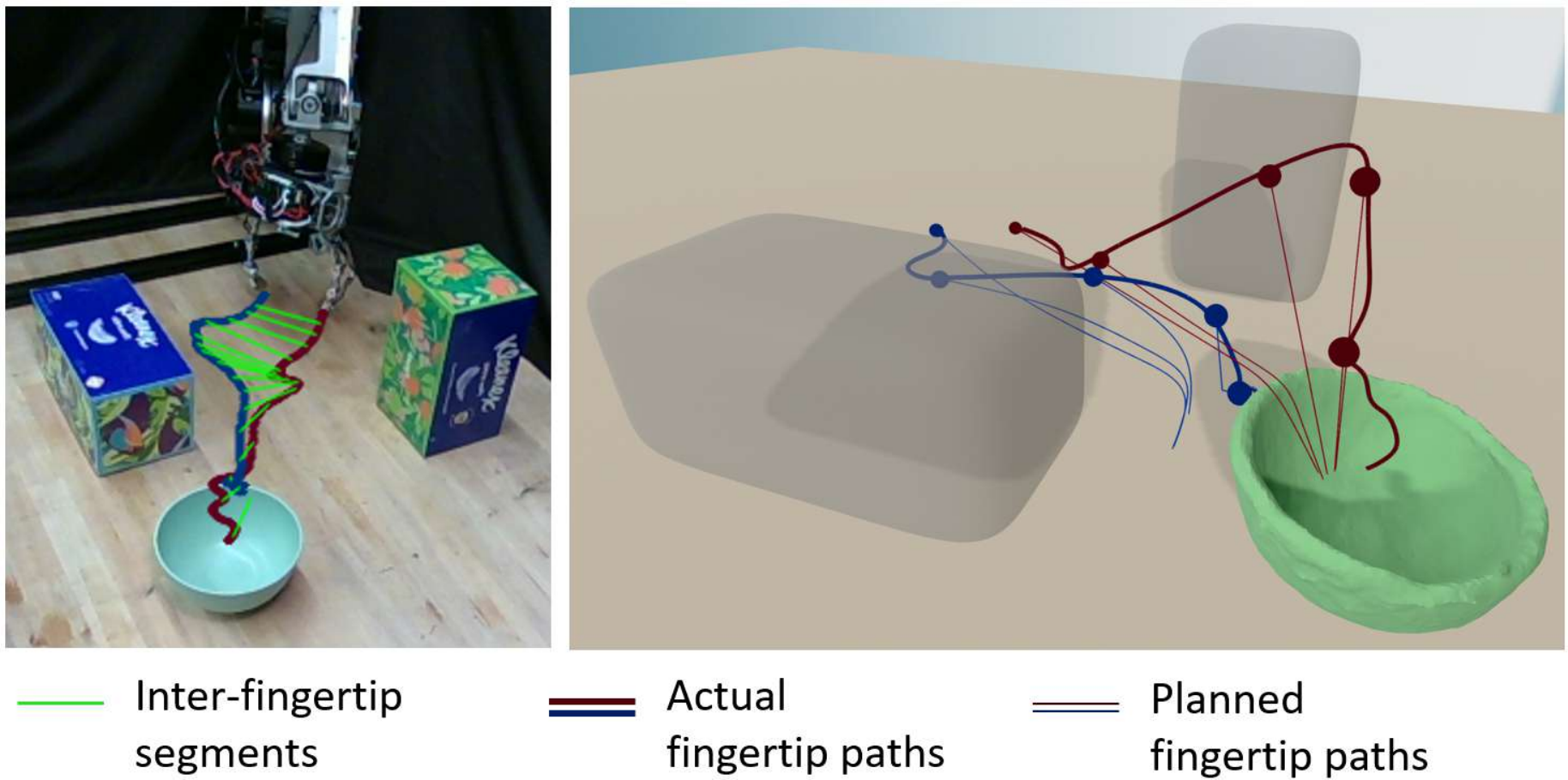}
    \caption{An example of successful navigation around obstacles. {\it Left}: Visualization of fingertip paths (blue and red) and instantaneous inter-fingertip segments (green). {\it Right}: Thicker curves show the executed fingertip paths, while thinner curves indicate the real-time planned fingertip paths.}
    \label{fig:bowl_two_obs_paths}
\end{figure}

Second, we evaluate grasping performance in the presence of obstacles, as illustrated in Fig.~\ref{fig:bowl_two_obs_paths}. The target object and obstacles are randomly placed, and the reaching controller is executed without external disturbances. Table~\ref{tab:success_rates_obstacles} reports the success rates over 10 trials for the cases with one, two, and three obstacles.
Fig.~\ref{fig:bowl_two_obs_paths} illustrates a successful case. The executed fingertip paths are shown with inter-fingertip segments ({\it Left}), alongside examples of planned fingertip paths ({\it Right}). From the right figure, two observations can be made: (i) the target grasp positions change as the motion progresses, and (ii) the planned and executed paths do not coincide exactly. The first observation is expected, since grasp targets are updated online based on the current fingertip positions. The second arises because the low-level joint-space QP must enforce constraints and track additional velocity objectives at the same time, making exact tracking of the planned fingertip velocities inherently impossible.
As shown in the figure, the planned fingertip paths nevertheless serve to continuously guide the fingertip motion toward the targets in real time, while avoiding collisions with the obstacles and the object.


\begin{figure}[!t]
    \centering
    \includegraphics[width=1\linewidth]{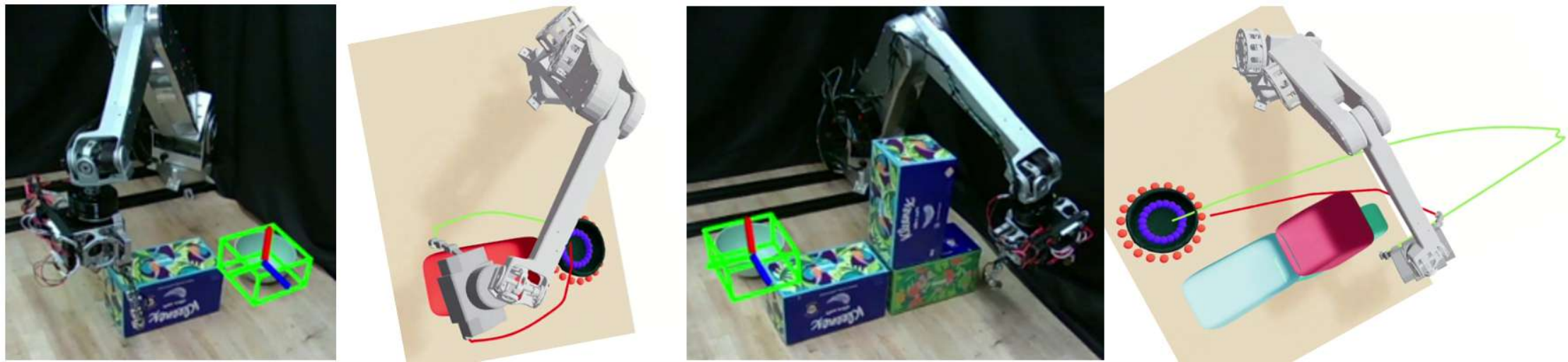}
    \caption{Failure cases due to obstacles and walls.}
    \label{fig:fail}
    \vspace{-10pt}
\end{figure}

Fig.~\ref{fig:fail} presents examples of failure cases, showing a limitation of our approach. In particular, the high-level velocity targets -- especially for fingertip positions -- are planned independently of the current full joint configuration. As a result, although the paths may appear collision-free from the fingertip’s perspective, following them can lead the robot into obstacle-facing configurations ({\it Left}) or local minima in narrow passages that it cannot traverse ({\it Right}). 

\section{Discussion and Conclusion}
We have proposed a hierarchical control framework for reactive grasping that is robust to external disturbances, where the higher layer defines task-space (or joint-subspace) velocity targets and the lower layer tracks them under full joint-level constraints with appropriate weighting. The framework enables robust grasping of highly concave objects, such as wine glasses and bowls, even in the presence of obstacles, thus advancing reactive manipulation in cluttered and dynamic environments.

Our framework can be improved in two directions. First, it relies on pre-defined grasp points, which limits adaptability to novel objects; this could be addressed by incorporating learning-based methods to predict grasp points online. Second, in cluttered environments with multiple obstacles, the system may become trapped in local minima when only fingertip velocity targets are considered (see Fig.~\ref{fig:fail}); adding velocity targets at the elbow or other joints, potentially through lower-frequency joint-level optimization as in~\cite{koptev2024reactive}, is a promising extension.







\bibliographystyle{IEEEtran}  
\bibliography{ref}

\end{document}